\documentclass[10pt,journal,letterpaper,twoside]{IEEEtran}

\usepackage{lineno}  
\usepackage{amsmath,amssymb,amsmath,bm}
\usepackage{color}
\usepackage{times}
\usepackage{multirow}
\usepackage{algorithm,algorithmic}
\usepackage{graphicx}
\usepackage{subfigure}
\usepackage{float}
\usepackage{hyphenat}

\newcommand{\eg}{e.g.}
\newcommand{\ie}{i.e.}

\newcommand{\etal}{et al.}

\renewcommand{\algorithmicrequire}{\textbf{Input:}}   
\renewcommand{\algorithmicensure}{\textbf{Output:}}  
\renewcommand{\bm}{\mathbf}
\DeclareFixedFont{\mf}{OT1}{ptm}{m}{n}{10pt}
\DeclareFixedFont{\mfb}{OT1}{ptm}{bx}{n}{10pt}
\begin{document}


\title{Exploring global diverse attention via pairwise temporal relation for video summarization}


\author{Ping~Li, Qinghao~Ye, Luming~Zhang, Luming~Zhang, Li~Yuan, Xianghua~Xu,
        and Ling~Shao
\thanks{This work was supported in part by the National Natural Science Foundation of China under Grants 61872122, 61502131, in part by Natural Science Foundation of Zhejiang Province under Grant LY18F020015. \emph{(Corresponding author: Ping Li.)}}
\thanks{P.~Li, Q.~Ye and X.~Xu are with the School of Computer Science and Technology, Hangzhou Dianzi University, Hangzhou 310018, China (e-mail: patriclouis.lee@gmail.com).}
\thanks{L.~Zhang is with the College of Computer Science, Zhejiang University, Hangzhou 310027, China.} 
\thanks{L.~Yuan is with the National University of Singapore, Singapore 119077.}
\thanks{L.~Shao is with Inception Institute of Artificial Intelligence, Abu Dhabi, UAE.}
}
\markboth{Draft}
{LI \MakeLowercase{\textit{et al.}}:~Exploring global diverse attention via pairwise temporal relation for video summarization}
%

\maketitle

\begin{abstract}
  Video summarization is an effective way to facilitate video searching and browsing. Most of existing systems employ encoder-decoder based recurrent neural networks, which fail to explicitly diversify the system-generated summary frames while requiring intensive computations. In this paper, we propose an efficient convolutional neural network architecture for video \textbf{SUM}marization via \textbf{G}lobal \textbf{D}iverse \textbf{A}ttention called \textbf{SUM-GDA}, which adapts attention mechanism in a global perspective to consider pairwise temporal relations of video frames. Particularly, the GDA module has two advantages: 1) it models the relations within paired frames as well as the relations among all pairs, thus capturing the global attention across all frames of one video; 2) it reflects the importance of each frame to the whole video, leading to diverse attention on these frames. Thus, SUM-GDA is beneficial for generating diverse frames to form satisfactory video summary. Extensive experiments on three data sets, \ie, SumMe, TVSum, and VTW, have demonstrated that SUM-GDA and its extension outperform other competing state-of-the-art methods with remarkable improvements. In addition, the proposed models can be run in parallel with significantly less computational costs, which helps the deployment in highly demanding applications.
\end{abstract}


 \ifCLASSOPTIONpeerreview
 \begin{center} \bfseries EDICS Category: 3-BBND \end{center}
 \fi


\section{Introduction}
\label{sec1:intro}
\IEEEPARstart{I}{n} the era of big data, video has become one of the most important carriers of data as the number and the volume have both increased rapidly in daily life. Video summarization as a good way to manage these videos (\eg, video searching and video browsing) \cite{Mei2015Video,li2017key} has received much interests in the field of computer vision and pattern recognition. It essentially selects the key shots of a video as the summary, and these key shots are expected to convey the most important information of video. To obtain the summary, traditional methods often use hand-crafted features which are then processed by unsupervised models \cite{yuan2019cycle-sum,Avila2011VSUMM} or supervised models \cite{Ghosh2012Discovering, gygli2014creating}. Nevertheless, these models can not be trained efficiently in an end-to-end manner and hand-crafted features are incapable of encoding high-level semantic information of video, so they may not work well for those videos with diverse and complex scenes in the real-world applications.

In the last decade, many contributions have been devoted to exploring the recurrent encoder-decoder architecture which utilizes Recurrent Neural Networks (RNNs) \cite{giles1994dynamic} and Long Short-Term Memory (LSTM) models~\cite{hochreiter1997long, Huang2019attention}. These models are able to learn high-level feature representation of data, thus generating video summary with high quality. Actually, recurrent neural networks can inherently capture temporal dependency by encoding sequence information of video frames, and enjoy the widespread success in practical tasks, \eg, machine translation~\cite{cho2014learning} and action recognition~\cite{donahue2016long}. However, there are several following drawbacks: (1) it is difficult for RNNs to make full use of GPU parallelization, due to the fact that the generation of its hidden state $h_t$ for the $t$-th frame depends on the previous hidden state $h_{t-1}$; (2) gated RNN models like LSTMs cannot well model long-range dependency across video frames since the gated mechanism will lead to serious decay of the history information inherited from those frames appearing at early time; (3) in some scenarios, \eg, the video news broadcast on one topic usually consists of several edited videos from different sources, there may exist semantic discontinuity within video sequences, which is very challenging and RNNs cannot well resolve this problem.

To alleviate the above problems, we exploit the temporal context relations among video frames from the pairwise relation perspective. In particular, a pairwise similarity matrix is designed to model multi-scale temporal relations across frames by storing the context information. The rationale behind this idea is that temporal relations of video frames can be modeled by the pairwise similarity between two frames regardless of their distance. That is to say, unlike RNNs that model long-range dependency using history information of early frames, it is unnecessary for our scheme to go through all the intermediate frames between source frame and target frame. Instead, we compute the pairwise similarity matrix directly at limited computational cost, which makes it quite efficient and suitable for GPU parallelization.

In another aspect, a good video summary should include those shots with the most diversity and the most representativeness, \ie, the selected key shots with higher importance scores should reflect diversified semantic information of video. Our human beings always tend to summarize video after scanning all the frames, which inspires us to imitate this process and summarize video by fully attending to the complete video as attention mechanism is prevailing and successful in sequence modeling \cite{vaswani2017attention}. Therefore, we develop an efficient video \textbf{SUM}marization model with \textbf{G}lobal \textbf{D}iverse \textbf{A}ttention called \textbf{SUM-GDA} using pairwise temporal relation, to quantify the importance of each video frame and simultaneously promote the diversity among these frames. In concrete, the proposed GDA mechanism is used to model the temporal relations among frames by leveraging the pairwise similarity matrix, where each row vector is closely related to the attention level and its lower entry indicates a higher dissimilarity weight. Hence, more emphasis should be put on the corresponding pairwise frames in order to promote the diversity among video frames. As a result, all stacked row vectors can reflet different attention levels of the whole video from a global perspective.

\begin{figure}[t]
\centering
\includegraphics[width=0.48\textwidth]{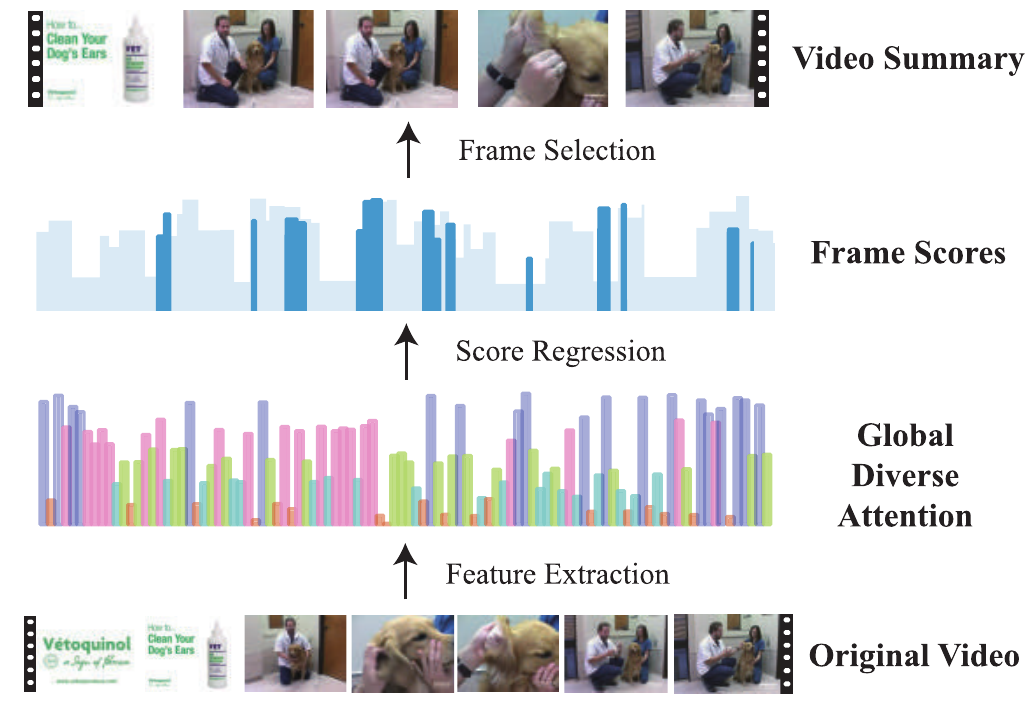}
\caption{The Overview of SUM-GDA. It exploits the global diverse attention mechanism to diversify the selected frames as the final summary. In Row3: The heights of colorized bars represent different measurements of frames, \ie, attention weights, which are computed as dissimilarity scores based on pairwise information within video. Such kind of different attention weights can reflect the diversity of frames in a global view. These diverse attention weights are utilized to predict the importance score of each frame in the video. In Row2: The light blue bars represent the predicted importance scores of video frames, and the dark blue bars indicate the frames selected for generating summary.}
\label{fig:overview}
\end{figure}

The overview of SUM-GDA is depicted in Figure~\ref{fig:overview}, where the key shots of input video are selected according to the frame scores that reflect the importance of the frame to the whole video. These scores are obtained by score regression with global diverse attention mechanism, which is the core component of the proposed model. In this work, we have explored both supervised and unsupervised variants of this model, which are evaluated by conducting a lot of interesting experiments on several video data sets including SumMe~\cite{gygli2014creating}, TVSum~\cite{song2015tvsum}, and VTW~\cite{zeng2016generation}, in different data settings, \ie, \textit{Canonical}, \textit{Augmented}, and \textit{Transfer}. Empirical studies demonstrate that the proposed method outperforms other state-of-the-art approaches in both supervised and unsupervised scenarios. Furthermore, the selected key shots for some randomly sampled videos are visually shown to further validate the effectiveness of the global diverse attention mechanism.

In short, the main contributions of this work can be highlighted in following aspects:
\begin{itemize}
    \item A global diverse attention mechanism is developed to model the temporal dependency of video by using pairwise relations between every two frames regardless of their stride magnitude, which helps well handle the long-range dependency problem of RNN models.

	\item By directly calculating the pairwise similarity matrix that reflects the relations between source frame and target frame, SUM-GDA only needs very
        limited computational costs so that it is inherently much more efficient than other competing approaches.

    \item The proposed SUM-GDA model is explored in supervised, unsupervised and semi-supervised scenarios. Empirical studies in terms of both quantitative and qualitative views are provided.

    \item The diversity of generated summaries and the influence of optical flow features are both investigated.
\end{itemize}

The remaining parts are organized as follows. Sec.~\ref{prior_work} generally reviews some closely related works and Sec.~\ref{model} introduces the global diverse attention mechanism and the SUM-GDA model in two scenarios. Then, Sec.~\ref{experiments} describes a number of experiments carried out on several data sets and reports both quantitative and qualitative results as well as rigorous analysis. Finally, we conclude this paper.

\begin{figure*}[t]
\centering
\includegraphics[width=0.95\textwidth]{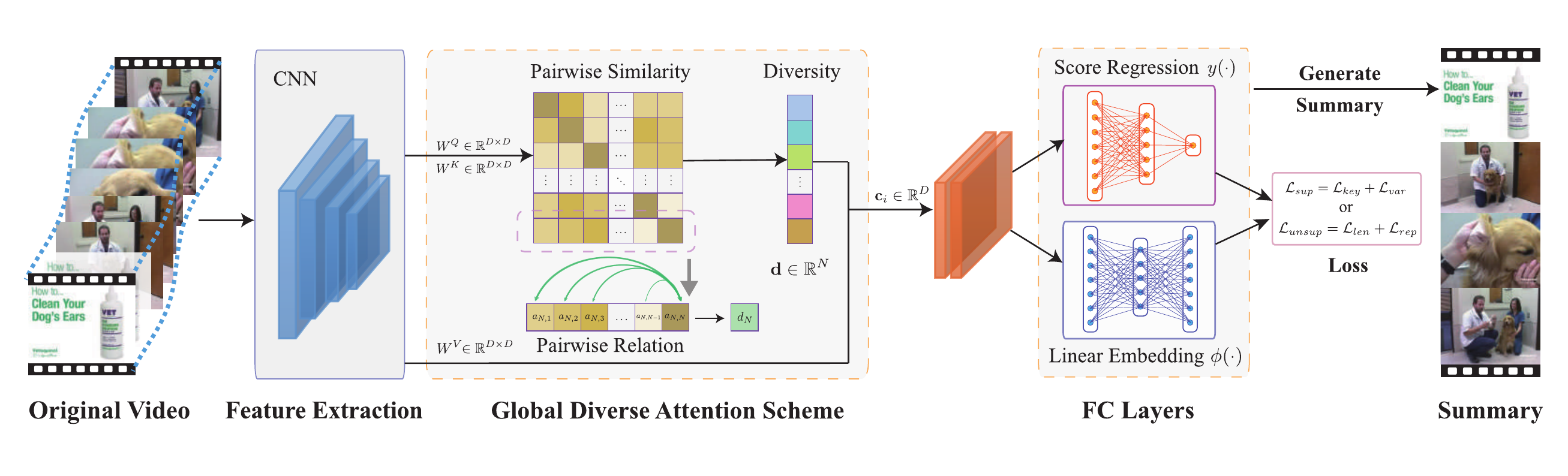} \vspace{-0.3in}
\caption{The architecture of SUM-GDA. Note that $W^Q$, $W^K$ and $W^V$ are three parameters to be learned by training model, $\bm{d}$ is normalized pairwise dissimilarity vector, and $\bm{c}_i$ indicates the context of frame $\bm{x}_i$.}
\label{fig:arch}
\end{figure*}
%
\section{Related Work}
\label{prior_work}
Video summarization \cite{fang2006fuzzy} has been a long-standing problem in multimedia analysis with great practical potential, and lots of relevant works have been explored in recent years.

Traditional approaches can be generally divided into two categories, \ie, unsupervised learning and supervised learning. \textit{Unsupervised} methods generally choose key shots in terms of heuristic criteria, such as relevance, representativeness, and diversity. Among them, cluster-based approaches~\cite{Mundur2006Keyframe} aggregate visually similar shots into the same group, and the obtained group centers will be selected as final summary. In earlier works, clustering algorithms are directly applied to video summarization~\cite{Zhuang2002Adaptive}, and later some researchers combine domain knowledge to improve performance \cite{Mundur2006Keyframe, Avila2011VSUMM}. Besides, dictionary learning~\cite{Mei2015Video} is another stream of unsupervised methods, and it finds key shots to build a dictionary as the representation of video in addition to preserving the local structure of data when necessary. \textit{Supervised} methods mainly utilize human-labeled summaries as training data to learn how human would summarize videos, which benefits obtaining impressive results. For instance, Ghosh~\etal~\cite{Ghosh2012Discovering} and Gygli~\etal~\cite{gygli2014creating} treat video summarization as a scoring problem which depends on the interestingness and the importance of video frames respectively, where the shots with higher scores are chosen to generate video summary. Lu and Grauman~\cite{Lu2013Story} proposed a story-driven model to tell story of an egocentric video. Furthermore, there are also some attempts to exploit auxiliary information such as web images and video categories to enhance the summarization performance.

Recently, deep learning approaches have attracted increasing interest for summarizing videos. For example, Yao~\etal~\cite{Yao2016Highlight} proposed a deep ranking model based on convolutional neural networks to encode the input video and output the ranking scores according to the relationship between highlight and non-highlight video segments; Zhang~\etal~\cite{zhang2016video} applied bidirectional LSTM to video summarization, and it can predict the probability of each selected shot; Zhang~\etal~\cite{zhang2016video} also developed DPP-LSTM, which further introduces Determinantal Point Process (DPP) to vsLSTM; Zhao~\etal~ \cite{Zhao2017Hierarchical} put forward a hierarchical architecture of LSTMs to model the long temporal dependency among video frames; Yuan~\etal~ \cite{yuan2019cycle-sum} designed cycle-consistent adversarial LSTM networks to sufficiently encode the temporal information in an unsupervised manner, resulting in promising summarization performance. In addition, Zhou~\etal~\cite{zhou2018deep} employed reinforcement learning and Rochan~\etal~ \cite{rochan2018video} adapted the convolutional sequence model to promote the quality of video summary; Rochan~\etal~ \cite{rochan2019video} attempted to train the model with unpaired samples by adopting key frame selector and summary discriminator network in unsupervised manner, and the generated summary may come from different source videos.

Inspired by human visual perception~\cite{ba2014multiple}, attention mechanism is exploited to guide the feed forward process of neural networks. To this end, attention-based models have gained satisfactory performance in various domains, such as person re-identification \cite{yang2019attention}, visual tracking \cite{Chen2019multiattention}, video question answering \cite{Wang2020vqa}, and action recognition \cite{li2020spatio}. Among them, self-attention \cite{vaswani2017attention} mechanism leverages the relation among all points of a single sequence to generate the corresponding representation of data, which makes it be widely applied in real-world tasks \cite{wang2018non}, \eg, Fajtl~\etal~ \cite{fajtl2018summarizing} utilized self-attention to only measure the importance of each frame while neglecting frame diversity. However, our work takes advantage of self-attention mechanism to model the temporal pairwise relation to quantify the importance of each frame and also to promote the diversity among the frames, leading to boosted summarization performance.

\section{The Proposed Approach}
\label{model}
In this section, we first introduce the proposed video summarization model with global diverse attention (SUM-GDA) in a supervised manner, which employs convolutional neural networks as backbone. The developed model is expected to be much more efficient compared to the typical recurrent neural networks like LSTMs, since existing RNNs based methods always have the difficulty in modeling the long-range dependency within videos especially in surveillance environment. To overcome this difficulty, we propose the global diverse attention mechanism by adapting self-attention \cite{vaswani2017attention} into video summarization task. In particular, we design a pairwise similarity matrix to accommodate diverse attention weights of video frames, which can well encode temporal relations between every two frames in a wide range of stride (\ie, the gap between source frame and target frame). Diverse attention weights are then transformed to importance scores for every frame, and importance score indicates the importance of the frame that may characterize some critical scene or objects within the given video.

Essentially, SUM-GDA leverages the global diverse attention mechanism to derive the dissimilarity representation of video frames, which makes it possible to accomplish video summarization even when the sequence order of video frames is disrupted in some situations, \eg, several videos regarding the same news are combined together by careful editing. Thus, SUM-GDA can be trained in parallel on multiple GPUs, which significantly reduces computational costs.

\subsection{Global Diverse Attention}

Global diverse attention mainly exploits diversity information to enhance the performance of video summarization. Particularly, GDA adopts self-attention to encode the pairwise temporal information within video frames, and then aggregates the pairwise information of all frame pairs in a video to globally evaluate the importance of each frame to the whole video. Afterwards, GDA transforms importance scores into pairwise dissimilarity scores which indicate the appearance variation between two frames.

Given a video $V$, it consists of $N$ frames $\{v_1, v_2, \ldots, v_N\}$. For its $i$-th frame and $j$-th frame, $i,j \in \{1, 2, \ldots, N\}$, the corresponding feature representation vectors $\bm{x}_i \in \mathbb{R}^D$ and $\bm{x}_j \in \mathbb{R}^D$ are derived from pre-trained CNN networks like GoogLeNet \cite{szegedy2015going}. The pairwise attention matrix $A \in \mathbb{R}^{N\times N}$ essentially reveals the underlying temporal relation across frame pairs of the video, and the entry $A_{ij}$ between the $i$-th frame  and the $j$-th frame is
\begin{equation}
\label{self-attention}
A_{ij} =\frac{1}{\sqrt{q}}(W^Q\bm{x}_i)^T(W^K\bm{x}_j),
\end{equation}
where $q>0$ is a constant; two linear projections $W^Q \in \mathbb{R}^{D\times D}$ and $W^K \in \mathbb{R}^{D\times D}$ correspond to the paired video frames $\bm{x}_i$ and $\bm{x}_j$ respectively, and they are parameters to be learned by training model. Instead of directly using dot-production, we scale the output attention value with an empirical scaling factor $\frac{1}{\sqrt{q}}$ ($q = D$), because the model is very likely to generate very small gradients after applying a softmax function without scaling especially during back-propagation process.

The obtained pairwise attention weights $A_{ij}$ are then converted to corresponding normalized weights $\alpha_{ij}$ by using following softmax function, \ie,
\begin{equation}
\label{softmax}
\alpha_{ij} = \frac{\exp(A_{ij})}{\sum_{r=1}^{N}\exp(A_{rj})}.
\end{equation}
These pairwise attention weights only reveal the importance of each frame to the video, while a good summary is expected to be composed by diverse frames or key shots. Hence, to diversify video frames globally, GDA attends to those frames that are dissimilar to the other frames in the whole video. Mathematically, we compute the pairwise dissimilarity vector $\hat{\bm{d}} = [\hat{d}_1, \hat{d}_2, \ldots, \hat{d}_N]\in \mathbb{R}^{N}$ by
\begin{equation}
\label{dissimilar}
\hat{d}_i = \prod\limits_{j=1}^N (1 - \alpha_{ij}), \quad \bm{d} = \frac{\hat{\bm{d}}}{\|\hat{\bm{d}}\|_1},
\end{equation}
where $\|\cdot\|_1$ denotes $\ell_1$-norm and the operator $\prod$ represents the product of multiple elements. The vector $\bm{d}$ denotes the normalized pairwise dissimilarity and its elements $\{d_i\}_{i=1}^N$ are actually global diverse attention weights, which indicate how the $i$-th frame $\bm{x}_i$ differs from the whole input video and simultaneously suggest the diversity of the frames.

To capture frame semantic information modeled by the weighted context matrix $C \in \mathbb{R}^{D\times N}$, we apply linear mapping to normalized pairwise dissimilarity vectors as
\begin{equation}
\label{weighted}
\bm{c}_i = d_i \otimes (W^V \bm{x}_i),
\end{equation}
where the linear projection matrix $W^V \in \mathbb{R}^{D\times D}$ is the parameter to be learned, $\bm{c}_i \in \mathbb{R}^D$ is the weighted vector reflecting the context of the $i$-th frame, and $\otimes$ is element-wise product.

\subsection{The SUM-GDA Model}
Incorporating with global diverse attention, the proposed SUM-GDA model includes the score regression module and the linear embedding module following the feed forward layer, and its architecture is illustrated in Figure~\ref{fig:arch}. The model first extracts feature vectors by pre-trained CNN and computes global diverse attention matrix $A$. Then, the weighted features are handled by two fully-connected layers including linear embedding function $\phi(\cdot)$ and score regression function $y(\cdot)$. Those frames with high regression scores are selected to form final summary. In the feed forward layer, for weighted context matrix $C$, linear transformation is performed with the dropout layer using layer normalization technique. For score regression, we compute frame scores $\bm{y}\in \mathbb{R}^N$ using two linear layers with ReLU activation function, dropout, and layer normalization in between.  

Admittedly, the proposed model can predict the likelihood of one video frame to be included in the final summary. To further diversify the selected frames, we adopt the Determinantal Point Process (DPP)~\cite{kulesza2012dpp} technique which defines the distribution that measures the negative correlation over all the subsets. DPP has been widely used in summarization methods~\cite{gong2014diverse} as it can promote diversity within the selected subsets.

Given the index set $\mathcal{Y} = \{1, 2, \ldots, N\}$, the positive semi-definite kernel matrix $L \in \mathbb{R}^{N\times N}$ is computed to represent the frame-level pairwise similarity, and the probability of a subset $\mathcal{Y}_{sub} \subseteq \mathcal{Y}$ is
\begin{equation}
\label{dpp}
P_L(Y = \mathcal{Y}_{sub}; L) = \frac{\det(L_{\mathcal{Y}_{sub}})}{\det(L + I_N)},
\end{equation}
where $Y$ is the random variable to take an index value, $I_N$ is a $N \times N$ identity matrix, and $\det(L + I_N)$ is a normalization constant. If two items within the same subset are similar, then the probability $P(\{i,j\} \subseteq \mathcal{Y}; L) = L_{ii}L_{jj} - L_{ij}^2$ will be close to zero, \ie, setting the probability of the subset to zero. Otherwise, the high probability indicates that the subset has high variation or diversity.

Inspired by quality-diversity decomposition \cite{kulesza2012dpp}, we enhance DPP with explicitly modeling the variation by defining the following kernel matrix $L$ as
\begin{equation}
\label{kernel}
L_{ij} = y_iy_j \Phi_{ij} = y_iy_j \exp(-\beta\|\bm{\phi}_i - \bm{\phi}_j \|_2^2),
\end{equation}
where the pairwise similarity between the paired frames $\bm{x}_i$ and $\bm{x}_j$ is derived from two linear transformations $\bm{\phi}_i$ and $\bm{\phi}_j$, whose output frame scores are $y_i$ and $y_j$.

\paragraph{Variation Loss}
Since DPP enforces the high diversity constraint on the selection of frame subsets, the redundancy of video summary can be reduced. And such diversity can be evaluated by the variation loss, \ie,
\begin{equation}
\label{var_loss}
\mathcal{L}_{var} = -\sum_{\mathcal{Y}_{sub} \subseteq \mathcal{Y}}\log P_L(Y = \mathcal{Y}_{sub}; L).
\end{equation}

\paragraph{Keyframe Loss}
In the supervised setting, we use ground-truth annotations of key frames $\hat{y}$ during training and define key frame loss formulated as
\begin{equation}
\label{key_loss}
\mathcal{L}_{key} = -\sum_{i = 1}^N \big(\hat{y}_i\log y_i + (1-\hat{y}_i)\log(1-y_i)\big).
\end{equation}

By combining Eq. (\ref{var_loss}) and Eq. (\ref{key_loss}), we can obtain the supervised loss of the proposed SUM-GDA model:
\begin{equation}
\label{loss}
\mathcal{L}_{sup} = \mathcal{L}_{key} + \mathcal{L}_{var}.
\end{equation}

During model training, the above loss functions are optimized iteratively. By incorporating DPP with score regression, we can build a unified end-to-end deep neural network architecture SUM-GDA for video summarization, and its training procedures are summarized in Algorithm \ref{alg:train}.
\begin{algorithm}[!t]
\begin{algorithmic}[1]
\caption{SUM-GDA Model Training}
\label{alg:train}

\REQUIRE ~~\\
Set of $M$ videos $\mathcal{V} = \{V_1, V_2, \cdots, V_M\}$, learning rate $\eta$.
\ENSURE ~~\\
Learned model parameters: $\Theta$.\\

\STATE Initialize all parameters denoted by $\Theta$ using Xavier.
\STATE Extract frame-level features $\{X_m\}_{m=1}^M$ for all videos.

\REPEAT
	\FOR{$m = 1$ \TO $M$}
		\STATE Use $X_m$ to calculate global pairwise attention matrix $A \in \mathbb{R}^{N\times N}$ using Eq.(\ref{self-attention}).
		\STATE Calculate the normalized pairwise dissimilarity vector $\hat{\bm{d}} \in \mathbb{R}^{N}$ by \mbox{Eqs.(\ref{softmax})(\ref{dissimilar})}.
		\STATE Compute weighted context vector $\bm{c}_i$ by Eq.(\ref{weighted}).
		\STATE Calculate frame score $y_i$ by score regression $y(\cdot)$.
		\STATE Obtain transformed feature vector $\bm{\phi}_i$ by linear embedding function $\phi(\cdot)$.
		\STATE Compute the loss $\mathcal{L}$ using Eq.(\ref{loss}) or Eq.(\ref{loss_unsup}).
		\STATE $\Theta \leftarrow \Theta - \eta \bigtriangledown_{\Theta} \mathcal{L}(\Theta)$.
	\ENDFOR
\UNTIL{convergence}

\RETURN $\Theta$.

\end{algorithmic}
\end{algorithm}

\subsection{Unsupervised SUM-GDA}
In many practical tasks, $\text{SUM-GDA}_{unsup}$ can learn from a set of untrimmed videos without supervision.

Generally speaking, it is difficult for different users who are asked to evaluate the same video to achieve the consensus on the final summary. Instead of using the oracle, we replace the key frame loss in Eq.~(\ref{key_loss}) with the length regularization $\mathcal{L}_{len}$ balanced by summary ratio $\sigma$ shown below:
\begin{equation}
\label{len_loss}
\mathcal{L}_{len} = \Big\|\frac{1}{N}\sum_{i=1}^Ny_i - \sigma\Big\|_2.
\end{equation}

Besides, in order to fully represent the diversity of the selected frames without summary annotations, we use the repelling loss $\mathcal{L}_{rep}$ \cite{zhao2016energy} to enhance the diversity among video frames. This loss is calculated as the mean value of pairwise similarities for all $N$ video frames, \ie,
\begin{equation}
\label{rep_loss}
\mathcal{L}_{rep} = \frac{1}{N(N-1)}\sum_{i}\sum_{i\neq j}\frac{\bm{\phi}_i^T \bm{\phi}_j}{\|\bm{\phi}_i\|_2 \|\bm{\phi}_j\|_2}.
\end{equation}

The merit of repelling loss is that a diverse subset will lead to the lower value of $\mathcal{L}_{rep}$. Now, we can obtain the unsupervised loss of $\text{SUM-GDA}_{unsup}$ as follows:
\begin{equation}
\label{loss_unsup}
\mathcal{L}_{unsup} = \mathcal{L}_{len} + \mathcal{L}_{rep}.
\end{equation}
$\text{SUM-GDA}_{unsup}$ has great potential in practice since it does not require ground-truth annotations which is hard to collect due to its expensive human labeling.

\subsection{Summary Generation}
\label{gen_summ}
We generate the final summary of an input video by selecting a set of key shots. Following \cite{zhang2016video}, we first generate a set of change points using Kernel Temporal Segmentation (KTS)~\cite{potapov2014category}, which indicates the key shot segments, and then constrain the summary length $l$ to the proportion of user summary length to the original video length. After that, we select key shots by the $0/1$ Knapsack algorithm which is formulated as
\begin{equation}
\label{greedy}
\max_{p_k}\,\,  \sum_{k=1}^T p_k s_k, \quad
s.t.~
\begin{cases}
\sum\limits_{k=1}^T p_k l_k \leq l,   \\
s_k = \frac{1}{l_k}\sum\limits_{t=1}^{l_k} y_{k}^t,  \\
p_k \in \{0, 1\}.
\end{cases}
\end{equation}
where $s_k$ represents the mean score of a specific key shot within $T$ key shots with length $l_k$, and the key shots with $p_k = 1$ are selected to generate the final summary. Algorithm \ref{alg:summarization} gives the detailed steps to generate video summary using the proposed SUM-GDA model.

\begin{algorithm}[!t]
\caption{Video Summary Generation}
\label{alg:summarization}
\begin{algorithmic}[1]

\REQUIRE ~~\\
Test video $V$ and model parameters $\Theta$.\\
	
\ENSURE ~~\\
Video summary $\bm{S}$. \\

\STATE Initialization: $\bm{S} \leftarrow \emptyset$.
\STATE Extract frame features $X = [\bm{x}_1, \bm{x}_2, \cdots, \bm{x}_N] \in \mathbb{R}^{D\times N}$ of $V$ via pre-trained CNN model.
\STATE Use KTS \cite{potapov2014category} to partition the test video into different segments $\{S_k\}_{k=1}^T$.
\STATE Compute global pairwise attention matrix $A \in \mathbb{R}^{N\times N}$ using Eq.(\ref{self-attention}).
\STATE Calculate the normalized dissimilarity vector $\hat{\bm{d}} \in \mathbb{R}^{N}$ by \mbox{Eqs.(\ref{softmax})(\ref{dissimilar})}.
\STATE Compute weighted feature $\bm{c}_i \in \mathbb{R}^{D}$ via Eq.(\ref{weighted}).
\STATE Calculate frame score $y_i$ for each context feature $\bm{c}_i$ by passing feed forward layer and score regression module.
\STATE Get $p_k$ for each shot $S_k$ generated by Eq.(\ref{greedy}).
\FORALL{\text{shot $S_k$ in video $V$}}
	\STATE \textbf{if} $p_k = 1$, \textbf{then} \\
	\STATE \quad \quad$\bm{S} \leftarrow$ $\bm{S} \cup \{S_k\}$.
\ENDFOR

\RETURN $\bm{S}$.

\end{algorithmic}
\end{algorithm}

\subsection{Summary Diversity Metric}
\label{diversity}
To examine the quality of generated summary, most existing methods adopt the popular F-score \cite{zhang2016summary, zhang2016video} metric which computes the overlap shots between generated summary and user summary. However, this criterion fails to reflect the diversity within generated summary and such diversity can be actually regarded as another way to evaluate the quality of generated summary. Hopefully, the key shots forming the summary should be diverse as much as possible. In this work, we propose one summary diversity metric ($\zeta$) to assess the diversity within generated summaries, \ie,
\begin{equation}
\label{diverse}
\zeta = \frac{1}{M\bullet T} \sum_{m=1}^M \sum_{i=1}^T \mathop{min}_{\bm{S}_q \in \bm{S}, q = 1, \ldots, |\bm{S}|} (\| X_m^{\bm{S}_q} - X_m^{S_i}\|_2),
\end{equation}
where $\|\cdot\|_2$ is $\ell_2$ norm, $min(\cdot)$ is a function to yield the minimum value; $M$ indicates the number of videos, $|\bm{S}|$ is the number of selected key shots in a video, $S$ is the set of all shots in a video, $\bm{S}$ is the set of all selected key shots in a video, $X_m^{\bm{S}_q} \in \mathbb{R}^{D}$ denotes shot feature vector obtained by averaging feature vectors across all frames in the shot $\bm{S}_q$ of video $m$ ($V_m$).

Summary diversity metric ($\zeta$) measures the average Euclidean distance between the video key shot and its nearest cluster center, \ie, key shot. The smaller $\zeta$ indicates higher diversity within generated summary. The rationale behind this is that the key shot can be regarded as the cluster center in clustering viewpoint. Hence, the more closely aggregated clusters suggest the more diversified clusters, thus the more diversified key shots. Usually, the selected key shots are more representative than those unselected ones.

\section{Experiments}
\label{experiments}
This section mainly explores the summarization performance of the proposed SUM-GDA model on several data sets. First, we give some statistics of data sets and describe the evaluation metrics as well as the experimental settings. Then, the implementation details will be given, which is followed by the reported results, the ablation study as well as further discussions on semi-supervised scenario, summary diversity and optical flow features.

\subsection{Data Sets}
We have evaluated different summarization methods on two benchmark data sets, \ie, SumMe~\cite{gygli2014creating} and TVSum~\cite{song2015tvsum}. Two additional data sets, \ie, Open Video Project (OVP)~\cite{de2011vsumm} and YouTube~\cite{de2011vsumm}, are used to augment and transfer the training set. SumMe and TVSum include both cases where the scene changes quickly or slowly which is challenging. SumMe consists of 25 user videos and different annotations. TVSum has 50 user videos and each video is annotated by 20 users with frame-level importance scores; OVP contains 50 videos while YouTube has 39 videos.

We also examined the performance on the VTW~\cite{zeng2016generation} database which is much larger and contains 2,529 videos collected from YouTube web site. The videos in VTW are mostly shorter than those in SumMe and TVSum. Each video is annotated with highlighted key shots.

\subsection{Evaluation Metric}
Following the protocols in \cite{zhang2016summary, zhang2016video}, the similarity of the generated summary is evaluated by measuring its agreement with user-annotated summary. We use the harmonic mean of precision and recall, \ie, \textit{F-score}, as the evaluation metric. The correct parts are the overlapped key shots between user annotations and generated summary. Formally, the precision is $P = \frac{overlap}{prediction}$ and the recall is $R = \frac{overlap}{user\ annotation}$. Then the \textit{F-score} is formulated as:
\begin{equation}
\label{f_score}
F = \frac{2 \times P \times R}{P + R} \times 100 \%.
\end{equation}

\subsection{Evaluation Settings}
For SumMe and TVSum, we evaluate the proposed model in three settings including Canonical (C), Augmented (A), and Transfer (T), and the details are shown in Table~\ref{table:setting}. The data set is divided into two parts: 80$\%$ for training and 20$\%$ for testing. We adopt five-fold cross validation and report the results by averaging the F-scores over five testing splits. Following the protocol in \cite{gygli2014creating, song2015tvsum}, we take the maximum of \textit{F-score} over different reference summaries for SumMe, and computed the average of \textit{F-score} over those summaries for TVSum. For VTW, we evaluate the model in canonical setting, where the data set is split to two parts including 2,000 videos for training and the rest for test in line with \cite{zhao2018hsa}.
\begin{table}[!t]
\centering
\caption{Evaluation settings for SumMe. To evaluate TVSum, we swap SumMe and TVSum.}
\label{table:setting}
\begin{tabular}{l|c|c}
\hline
Setting    & Training                           & Testing    \\ \hline
Canonical  & 80\% SumMe                         & 20\% SumMe \\ \hline
Augmented  & \begin{tabular}[c]{@{}c@{}}80\% SumMe + OVP \\ + YouTube + TVSum\end{tabular} & 20\% SumMe \\ \hline
Transfer   & OVP + YouTube + TVSum              & SumMe      \\ \hline
\end{tabular}
\end{table}

\subsection{Implementation Details}
For fair comparison, we use the \textit{pool5} features of GoogLeNet pre-trained on ImageNet, as the feature vector $\bm{x}\in \mathbb{R}^{1024}$ for each frame. The number of hidden units is set to 1024, the dropout rate is set to 0.6, and the $L_2$ weight decay coefficient is set to $10^{-5}$. For $\text{SUM-GDA}_{unsup}$, $\sigma$ is $0.3$. We train our model using the Adam optimizer with the initial learning rate $5\times 10^{-5}$ for SumMe, $10^{-4}$ for TVSum and $5\times 10^{-4}$ for VTW, and training is terminated after $200$ epochs. All experiments were conducted on a machine with NVIDIA GTX 1080Ti GPU using PyTorch platform. 

\subsection{Quantitative Results}
We compare the proposed model with a number of state-of-the-art video summarization methods including both supervised and unsupervised approaches, for which we directly use the records reported in their original papers. The results are reported in Table~\ref{tab:CAT} and Table~\ref{tab:CATunsup} respectively for supervised and unsupervised methods.

\begin{table}[!t]
\begin{center}
\caption{Performance comparison (F-score \%) with supervised methods on SumMe and TVSum.}
\label{tab:CAT}
\resizebox{0.9\linewidth}{!}{%
\begin{tabular}{l|ccc|ccc}
\hline
\multirow{2}{*}{Method}   &      & SumMe &      &      & \multicolumn{1}{l}{TVSum} & \multicolumn{1}{l}{} \\ \cline{2-7}
                          & C    & A     & T    & C    & A                         & T                    \\ \hline
Bi-LSTM~\cite{zhang2016video}                      & 37.6 & 41.6  & 40.7 & 54.2 & 57.9                      & 56.9                 \\
DPP-LSTM~\cite{zhang2016video}                     & 38.6 & 42.9  & 41.8 & 54.7 & 59.6                      & 58.7                 \\
SUM-GAN$_{sup}$~\cite{mahasseni2017unsupervised}   & 41.7 & 43.6  & -    & 56.3 & 61.2                      & -                    \\
DR-DSN$_{sup}$~\cite{zhou2018deep}    & 42.1 & 43.9  & 42.6 & 58.1 & 59.8                      & 58.9                 \\
SUM-FCN~\cite{rochan2018video}                      & 47.5 & 51.1  & 44.1 & 56.8 & 59.2                      & 58.2                 \\
HSA-RNN~\cite{zhao2018hsa}                      & - & 44.1  & - & - & 59.8                      & -                 \\
CSNet$_{sup}$~\cite{jung2019discriminative}     & 48.6 & 48.7  & 44.1 & 58.5 & 57.1                      & 57.4                 \\
VASNet~\cite{fajtl2018summarizing}       & 49.7 & 51.1  & -    & \textbf{61.4} & \textbf{62.4}                      & -                \\
M-AVS~\cite{ji2020video}                 & 44.4 & 46.1  & -    & 61.0 & 61.8                      & -                 \\\hline \hline
SUM-GDA                      & \textbf{52.8} & \textbf{54.4}  & \textbf{46.9} & 58.9 & 60.1         & \textbf{59.0}                 \\ \hline
\end{tabular}
}
\end{center}
\end{table}

\begin{table}[!t]
\begin{center}
\caption{Performance comparison (F-score \%) with unsupervised methods on SumMe and TVSum.}
\label{tab:CATunsup}
\resizebox{0.9\linewidth}{!}{%
\begin{tabular}{l|ccc|ccc}
\hline
\multirow{2}{*}{Method}     &      & SumMe &      &      & \multicolumn{1}{l}{TVSum} & \multicolumn{1}{l}{} \\  \cline{2-7}
                            & C    & A     & T    & C    & A                         & T                    \\ \hline
SUM-GAN$_{rep}$~\cite{mahasseni2017unsupervised}   & 38.5 & 42.5  & -    & 51.9 & 59.3                      & -                    \\
SUM-GAN$_{dpp}$~\cite{mahasseni2017unsupervised}   & 39.1 & 43.4  & -    & 51.7 & 59.5                      & -                    \\
DR-DSN~\cite{zhou2018deep}                       & 41.4 & 42.8  & 42.4 & 57.6 &	58.4 					  &	57.8                 \\
CSNet~\cite{jung2019discriminative}   & \textbf{51.3} & \textbf{52.1}  & 45.1 & 58.8 & 59.0            & \textbf{59.2}                 \\
Cycle-SUM~\cite{yuan2019cycle-sum}     & 41.9 & -  & -     & 57.6 & -                      & -                 \\
UnpairedVSN~\cite{rochan2019video}     & 47.5 & -  & -     & 55.6 & -                      & -                 \\\hline \hline
$\text{SUM-GDA}_{unsup}$ & 50.0 & 50.2  & \textbf{46.3} & \textbf{59.6} & \textbf{60.5}                      &  58.8                    \\ \hline
\end{tabular}
}
\end{center}
\end{table}

On SumMe and TVSum, the compared supervised methods contain Bi-LSTM \cite{zhang2016video}, DPP-LSTM~\cite{zhang2016video}, SUM-GAN$_{sup}$ \cite{mahasseni2017unsupervised}, DR-DSN$_{sup}$ \cite{zhou2018deep}, SUM-FCN \cite{rochan2018video}, HSA-RNN \cite{zhao2018hsa}, CSNet$_{sup}$~\cite{jung2019discriminative}, VASNet~\cite{fajtl2018summarizing}, and M-AVS~\cite{ji2020video}; the compared unsupervised methods include SUM-GAN$_{rep}$/SUM-GAN$_{dpp}$ \cite{mahasseni2017unsupervised}, DR-DSN \cite{zhou2018deep}, CSNet \cite{jung2019discriminative}, Cycle-SUM \cite{yuan2019cycle-sum}, and UnpairedVSN~\cite{rochan2019video}. Bi-LSTM and DPP-LSTM employ LSTMs to encode the temporal information of video. Meanwhile, SUM-GAN and its variants that adopt generative adversarial networks utilize the generated summary to reconstruct video frames, and Cycle-SUM additionally uses reconstructed video to construct summary to further promote the performance. Moreover, reinforcement learning is introduced in DR-DSN to solve the summarization problem; SUM-FCN uses a fully convolutional sequence network to encode the information; HSA-RNN utilizes hierarchical structure RNN to model the information of different levels; CSNet exploits different strides and chunks to model the temporal relation; VASNet leverages self-attention to only measure the importance of each frame; M-AVS adopts the encoder-decoder network with attention; UnpariedVSN trains the model with unpaired data via key frame selector network and summary discriminator network.

On VTW, the compared methods are all supervised methods, since to our best knowledge there are no works using unsupervised method on this database. The compared approaches contain HD-VS \cite{Yao2016Highlight}, DPP-LSTM~\cite{zhang2016video}, and HSA-RNN \cite{zhao2018hsa}. Among them, HD-VS uses two-stream CNNs to summarize the video. The results are reported in Table~\ref{table:VTW}.
\begin{table}[!t]
\begin{center}
\caption{Performance comparison (F-score \%) on VTW.}
\label{table:VTW}
\begin{tabular}{l|c|c|c}
\hline
Method & Precision & Recall & F-score \\ \hline
HD-VS~\cite{Yao2016Highlight}   & 39.2     & 48.3  & 43.3   \\ \hline
DPP-LSTM~\cite{zhang2016video} & 39.7     & 49.5  & 44.3   \\ \hline
HSA-RNN~\cite{zhao2018hsa} & 44.3     & \textbf{54.8}  & 49.1   \\ \hline \hline
$\text{SUM-GDA}_{unsup}$ & 47.8     & 48.6  & 47.9  \\ \hline
SUM-GDA & \textbf{50.1}     & 50.7  & \textbf{50.2}  \\ \hline

\end{tabular}
\end{center}
\end{table}

From Table~\ref{tab:CAT}, \ref{tab:CATunsup} and \ref{table:VTW}, several interesting observations can be found as follows.
\begin{itemize}
\item SUM-GDA outperforms most of the competing models by large margins in different data settings. Our model is better than the rest by at least $2.8\%$ on SumMe, \ie, boosted by $3.1\%$, $3.3\%$ and $2.8\%$ in canonical, augmented and transfer settings respectively. Moreover, the unsupervised model $\text{SUM-GDA}_{unsup}$ yields the best performance by $1.2\%$ higher than the second-best one in transfer setting on SumMe while it gains the top performance on TVSum in both canonical and augmented settings. This has well validated the advantages of the developed GDA component which models the long-range temporal relations among video frames and those selected frames are further diversified by our framework.

\item SUM-GDA achieves larger performance improvement than that of $\text{SUM-GDA}_{unsup}$ on SumMe, but slightly smaller than that on TVSum. This might be due to the fact that SumMe is more challenging and it adopts the highest \textit{F-score} among several users which shows more concentration when doing evaluation. Therefore, using supervised approaches tends to make the prediction be close to the specific ground truth. However, the evaluation on TVSum averages \textit{F-scores} among several users, and the ground truth will be possibly misled as different users may not reach concesus. the consistent agreement. In consequence, unsupervised $\text{SUM-GDA}_{unsup}$ model will provide a step closer to the user summaries.

\item The proposed SUM-GDA obtains larger gains on VTW compared to other methods because global diverse attention exploits the temporal relations among all frame pairs and can better encode the importance of each frame.

\end{itemize}

\begin{figure}[t]
\centering
\includegraphics[width=0.48\textwidth]{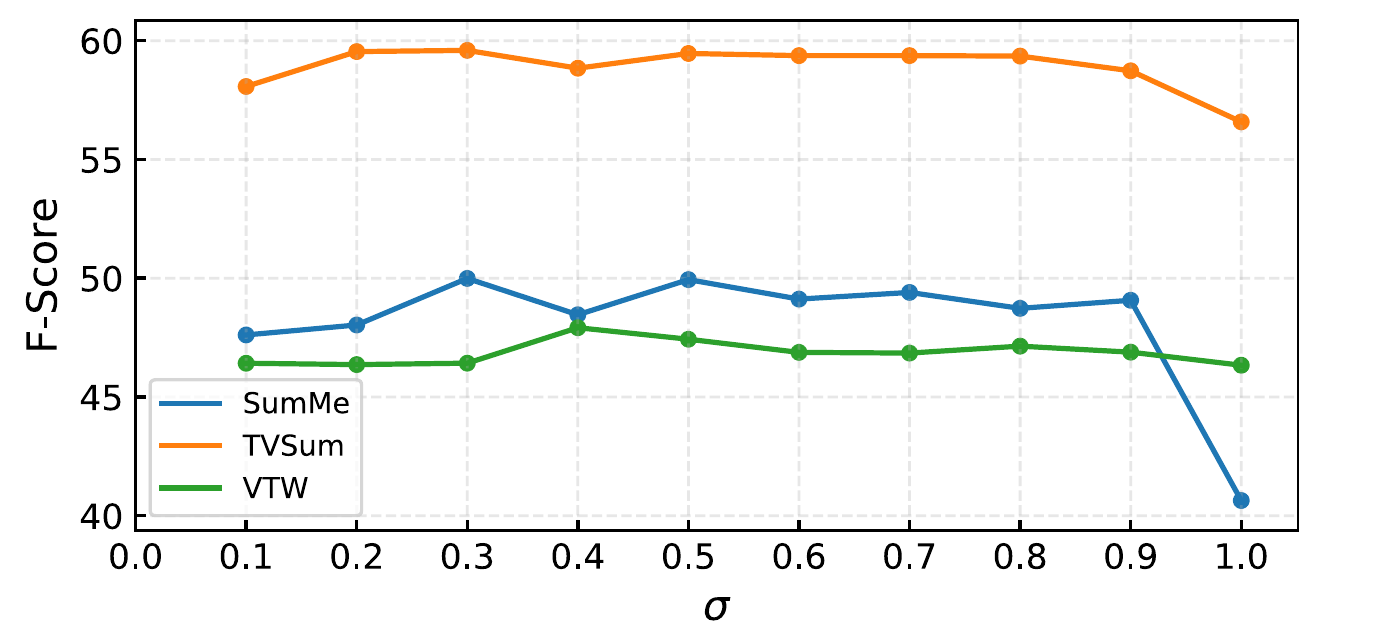}
\caption{F-score(\%) of $\text{SUM-GDA}_{unsup}$ for different summary ratios $\sigma$ on SumMe, TVSum, and VTW.}
\label{fig:sigma}

\end{figure}

Moreover, we have examined the sensitiveness of summary ratio $\sigma$ for SUM-GDA$_{unsup}$ to the summarization performance. Figure~\ref{fig:sigma} shows \textit{F-score} on all three databases. We can observe that when $\sigma$ is too large or too small, the performance degrades sharply. The best results are obtained when $\sigma = 0.3$. In addition, due to the large variance in summary proportion (\ie, the ratio of the summary length to the video length) on VTW, SUM-GDA$_{unsup}$ enjoys robust performance for $\sigma$. As shown by Figure~\ref{fig:sigma}, our model achieves the best when $\sigma = 0.4$, which is close to the mean value of all summary proportions on VTW.

\begin{figure}[t]
\centering
\includegraphics[width=0.48\textwidth]{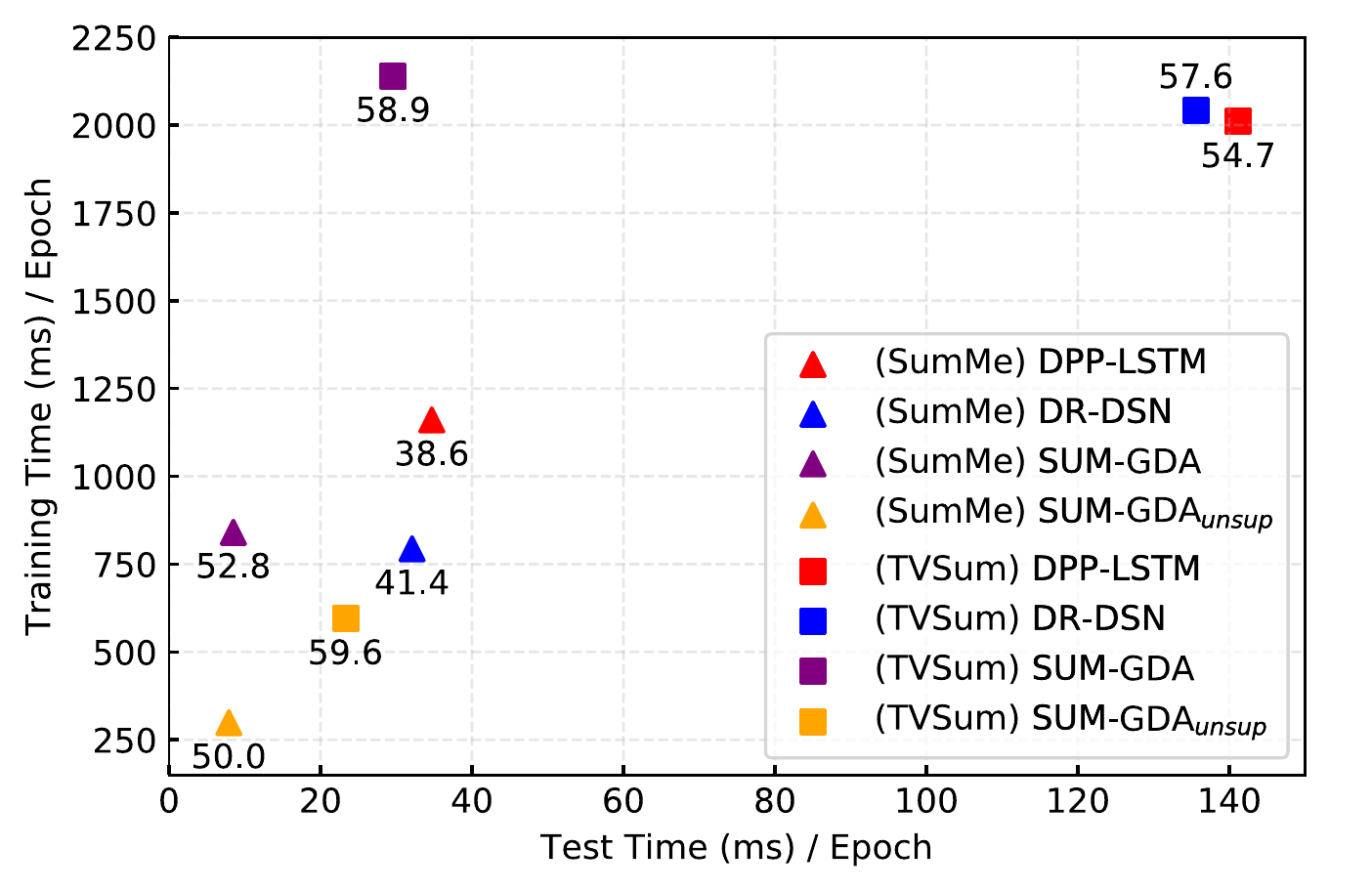}
\caption{Time comparison in terms of training and test time with F-score(\%) on SumMe and TVSum in Canonical setting.}
\label{fig:time}
\end{figure}

\subsection{Computational Issues}
We provide both training time and test time comparison for several state-of-the-art methods on SumMe and TVSum data sets in Figure~\ref{fig:time}, which shows SUM-GDA and $\text{SUM-GDA}_{unsup}$ are much more efficient. We found that $\text{SUM-GDA}_{unsup}$ reduces both training and test time significantly, which verifies that $\text{SUM-GDA}_{unsup}$ achieves better trade-off between efficiency and performance than existing models. In the following, we make some analysis on computational costs of the proposed method and RNN based methods.

Given an input sequence $\{\bm{z}_1, \cdots, \bm{z}_n\}$ which includes $n$ frame images with $\bm{z}_i \in \mathbb{R}^d$, suppose the dimension of the hidden state vector $\bm{h}_t$ is $d$. Then the time complexity of RNN is $O(n d^2)$ since $\bm{h}_t = tanh(\bm{W}_{hh} \bm{h}_{t-1} + \bm{W}_{zh} \bm{z}_t)$, where $\bm{W}_{xh}, \bm{W}_{hh} \in \mathbb{R}^{d\times d}$ \cite{li2018independently}. However, the GDA module of our approach only requires $O(n^2 d)$, which is faster than RNN when the sequence length $n$ is smaller than $d$. This is mostly true for frame image representations used by state-of-the-art video summarization models \cite{zhang2016video, zhou2018deep, yuan2019cycle-sum} who sample frames with $2fps$ from video. Regarding the benchmark databases SumMe, TVSum, and VTW tested above, the length of video is usually smaller than the dimension of its embedding representation, \eg, $1024$. Moreover, RNNs \cite{giles1994dynamic} and LSTM models~\cite{hochreiter1997long} suffer from the gradient vanishing and exploding problem due to the use of hyperbolic tangent and sigmoid activation function, which leads to gradient decay over time steps during the training process. Li \etal \cite{li2018independently} found that RNN and LSTM cannot keep the long-term temporal information of the sequence when its length is greater than $1000$ in empirical studies. In addition, the parallelized amount of computations can be measured by the minimum number of sequentially executed operations of the module \cite{vaswani2017attention}, since sequentially executed operation cannot be implemented in parallelization. For RNN based methods, $O(n)$ sequential operations are required while GDA module only needs $O(1)$ sequential operations if enough GPU cards are available. This implies that our GDA module is more efficient by taking advantages of parallelization. Hence, SUM-GDA is actually much faster than recurrent neural network based methods.

\subsection{Ablation Study}
To examine the influences of different loss terms, we conducted ablation study on the proposed model and the results are shown in Table~\ref{table:ablation}.

\begin{table}[!t]
\begin{center}
\caption{F-score (\%) of all cases on SumMe and TVSum in Canonical setting.}
\label{table:ablation}
\begin{tabular}{l|c|c}
\hline
Method     & SumMe & TVSum \\ \hline
SUM-GDA$_{unsup}$ w/o $ \mathcal{L}_{len}$ & 47.4  & 57.7  \\ \hline
SUM-GDA$_{unsup}$ w/o $ \mathcal{L}_{rep}$ & 48.9  & 58.9  \\ \hline
SUM-GDA$_{unsup}$ w/o GDA & 43.0  & 55.7 \\ \hline
SUM-GDA$_{unsup}$     & \textbf{50.0}  & \textbf{59.6}  \\ \hline \hline
SUM-GDA w/o $ \mathcal{L}_{var}$ & 51.6  & 58.1  \\ \hline
SUM-GDA w/o GDA	    & 45.1  & 53.1 \\ \hline
SUM-GDA     & \textbf{52.8}  & \textbf{58.9}  \\ \hline
\end{tabular}
\end{center}
\end{table}

As can be seen from the table, the model without length regularization (Row 1) will deteriorate the summarization performance, as the length of generated summary should be constrained in a sensible range. Besides, the performance of the model with repelling loss (Row 4) is improved by 1.1\% and 0.7\% on SumMe and TVSum respectively in comparison with that without such loss (Row 2). We can attribute this to the fact that when the repelling loss is added to the unsupervised extension besides length regularization, pairwise similarity between partial frames will be reduced, which would lower the importance scores of those frames in the video. Hence, the global diverse attention on video frames will be enhanced in some degree. Furthermore, the model with variation loss \ie SUM-GDA (Row 7) is better than that without it by 1.2\% and 0.8\% on SumMe and TVSum respectively, which is for the reason that variation loss helps generate diverse subsets. In addition, both of SUM-GDA$_{unsup}$ and SUM-GDA are improved by adopting GDA module as shown in Row 3 and Row 6. Therefore, different regularization terms or loss play different roles in promoting global diverse attention on video frames.

\subsection{Qualitative Results}
\begin{figure*}[!t]
\begin{center}
\def\arraystretch{0.9}
\begin{tabular}{@{}c@{\hskip 0.01\linewidth}c@{}}
    \includegraphics[width=0.48\linewidth]{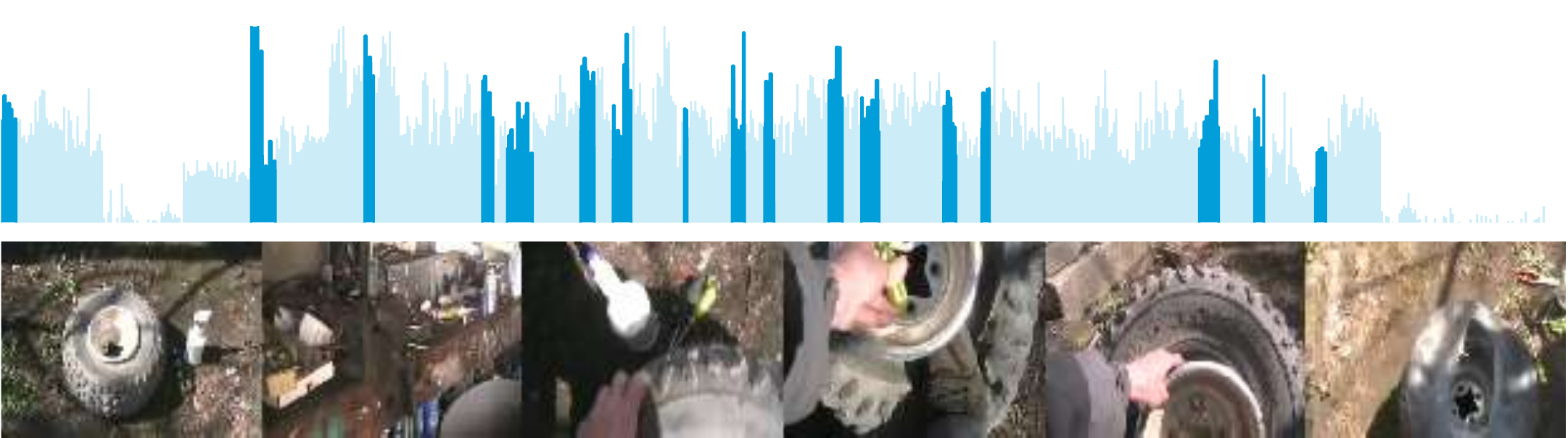}&
    \includegraphics[width=0.48\linewidth]{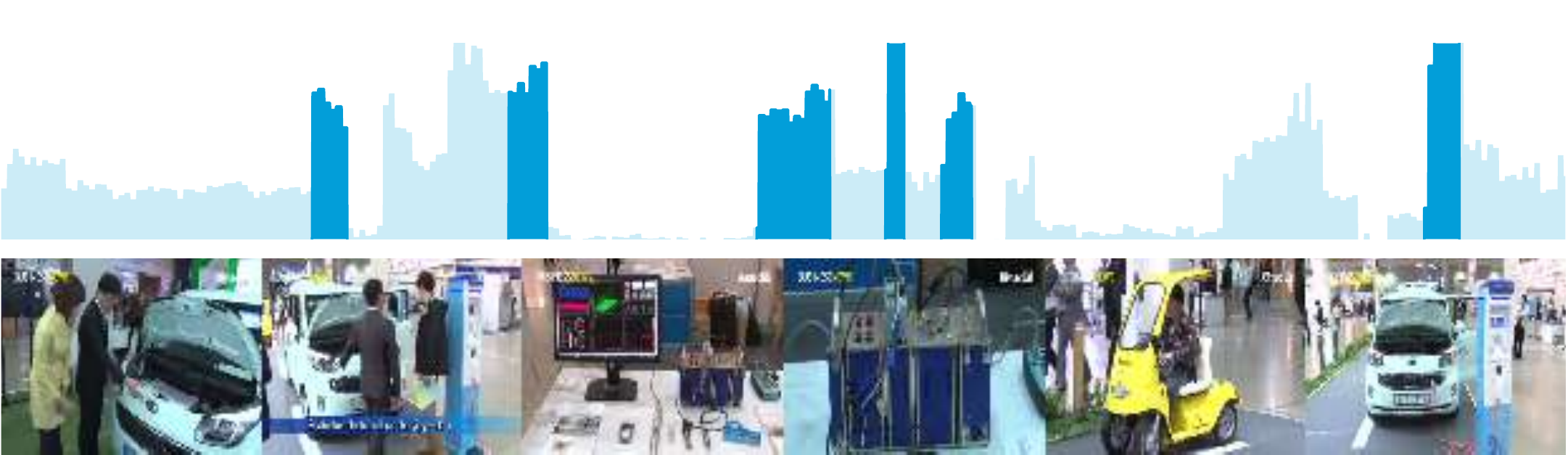}\\
{(a) Video 1: changing tire} & {(b) Video 10: production workshop}\\

    \includegraphics[width=0.48\linewidth]{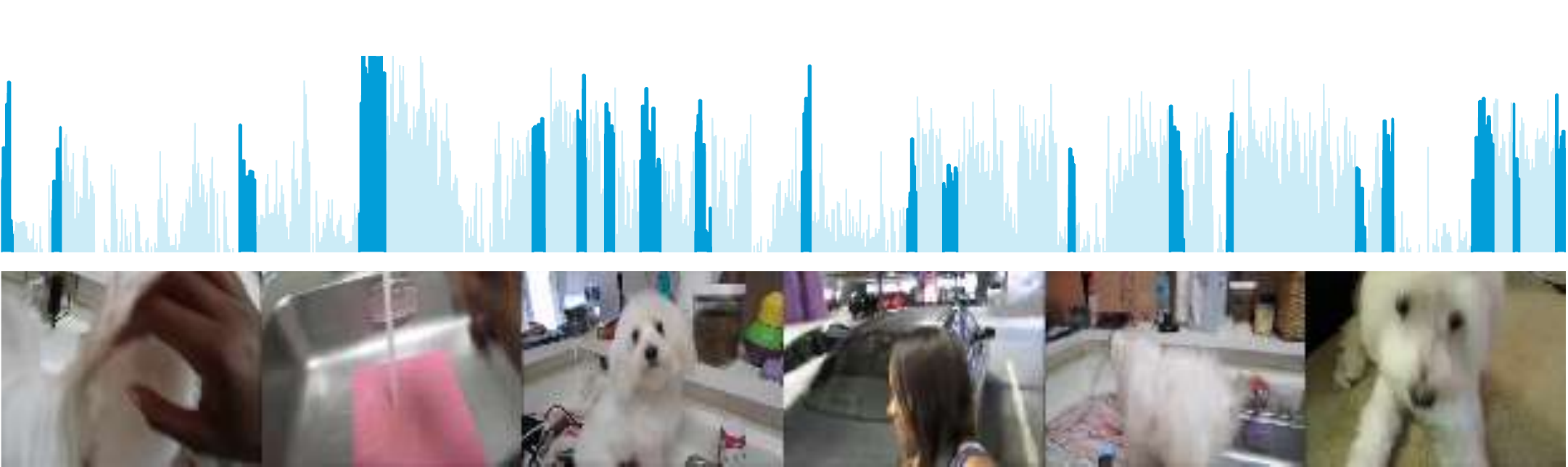}&
    \includegraphics[width=0.48\linewidth]{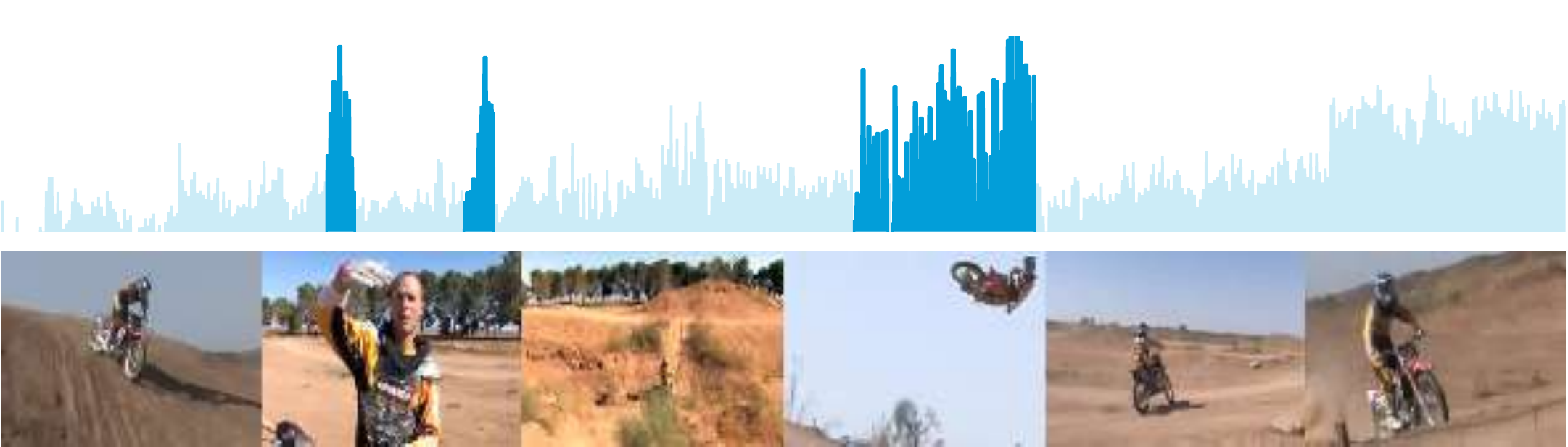}\\
{(c) Video 12: daily life of one woman} & {(d) Video 41: motorcycle flipping}
\end{tabular}
\end{center}
\caption{Visualization of SUM-GDA generated summaries for different videos in TVSum. Light blue bars represent ground-truth scores, and dark blue bars denote generated summaries.}
\label{fig:visual}
\end{figure*}

\begin{figure}[!t]
\begin{center}
\def\arraystretch{0.9}

\includegraphics[width=0.95\linewidth]{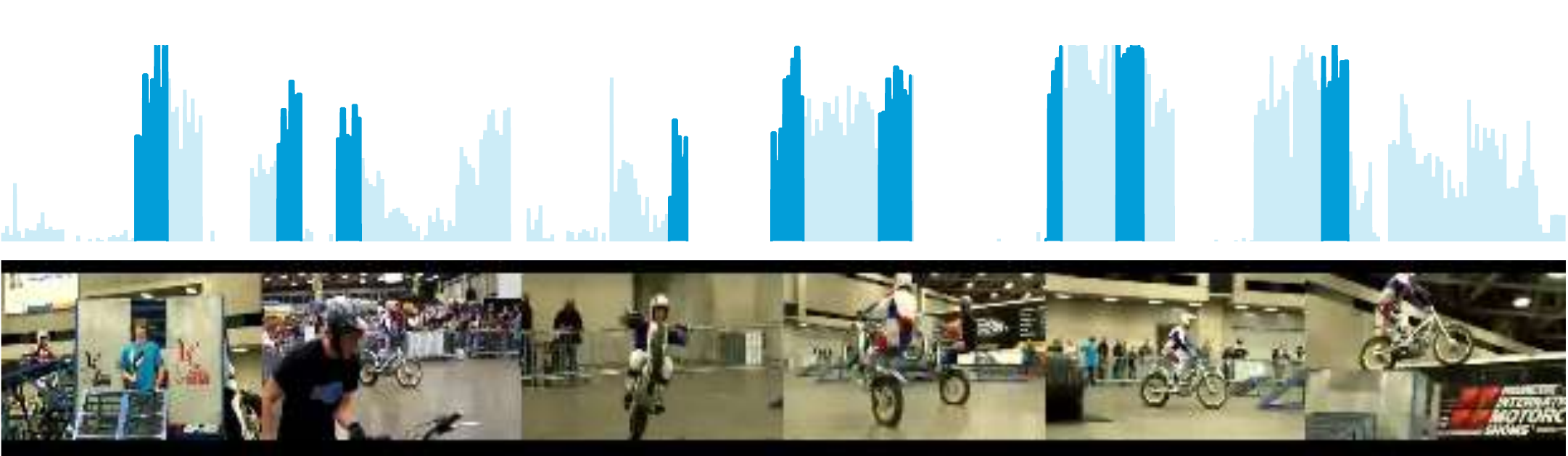}\\
{(a) SUM-GDA}\\

\includegraphics[width=0.95\linewidth]{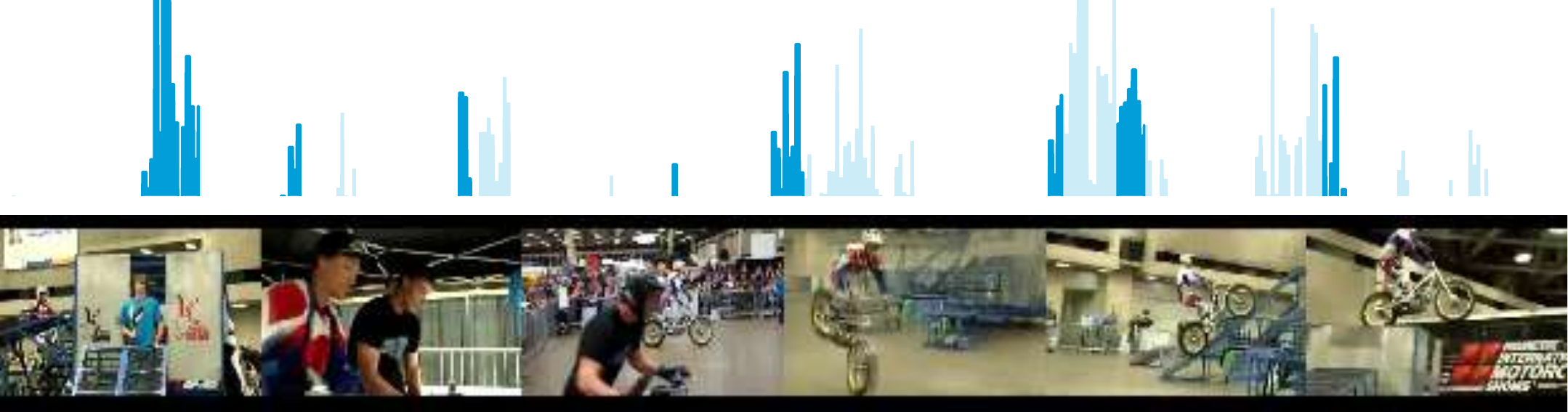}\\
{(b) $\text{SUM-GDA}_{unsup}$}
\end{center}
\caption{Visualization of selected key-shots for test video 30 (motorcycle show) in TVSum.}
\label{fig:sup_vs_un}
\end{figure}

We visualize the selected key shots of different videos on TVSum generated by our SUM-GDA model. From Figure~\ref{fig:visual}, it can be clearly seen that SUM-GDA selects most peak points according to frame scores using the $0/1$ Knapsack algorithm. As depicted by Figure~\ref{fig:sup_vs_un}, SUM-GDA and SUM-GDA$_{unsup}$ yield different frame scores. The bottom figure for SUM-GDA$_{unsup}$ generates frame scores with more sparsity, as we constrain the mean of frame scores to be close to the summary ratio, which may lead to sparsity.

\begin{figure*}[!t]
\begin{center}
\def\arraystretch{0.9}
\includegraphics[width=1.0\linewidth]{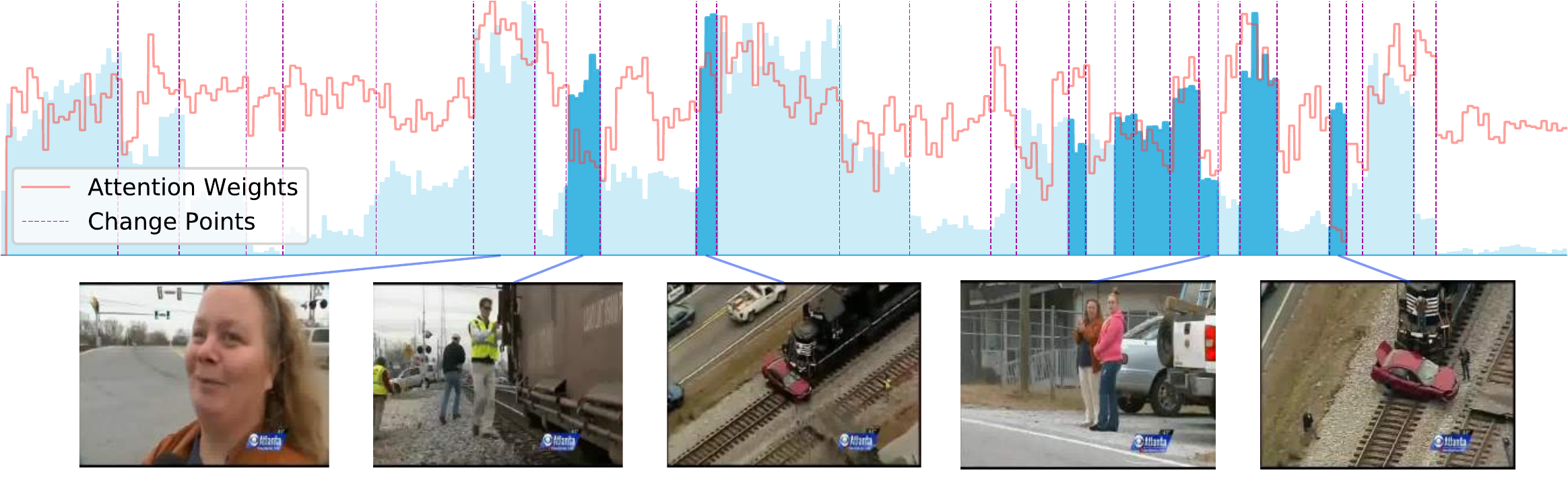}
\end{center}
\caption{Visualization of frame scores and attention weights on Video 7 in TVSum by SUM-GDA. Light blue bars represent predicted scores, and dark blue bars denote generated summaries. The magnitude of attention weights which is normalized to the same range of corresponding scores is plotted with light red line. Purple dash lines are the change points generated by KTS~\cite{potapov2014category}.}
\label{fig:attention}
\end{figure*}

To verify the effectiveness of SUM-GDA, we visualize global diverse attention weights on TVSum in Figure~\ref{fig:attention}, which describes a traffic accident where a train crashed with a car. Comparing the middle image and the right-most image, they look visually similar but their attention weights are very different. This does help achieve the goal of diversifying the selected frames because there exists redundancy in similar frames, only one of which is required to form the final summary in practice.

\subsection{Semi-Supervised Scenario}
In many practical applications, it often appears that only partial videos have labels due to costly human labeling while a large number of unlabeled videos are easily available which are useful for training summarization model. Regarding this scenario, we additionally examined the proposed methods in semi-supervised setting (which are not explored in previous works). Specifically, for SumMe, we use 80\% SumMe + OVP + YouTube + TVSum for training (labeled videos) and 20\% SumMe for testing (test videos). For TVSum, the above SumMe and TVSum are swapped. Those unlabeled videos are sampled from VTW by ignoring the corresponding annotations. Here, test videos are only used during test, which is slightly different from transductive learning which considers the test videos as unlabeled samples used for learning model. %

The results on SumMe and TVSum are shown in Table~\ref{tab:semi}, from which it can be observed that the summarization performance is consistently improved with the increasing number of unlabeled samples. This indicates that unlabeled data can also benefit the model learning and more unlabeled data can provide more consolidated information to enhance the quality of generated summaries.
\begin{table}[!t]
\begin{center}
\caption{Performance (F-score \%) with different numbers of unlabeled samples in semi-supervised setting. The first row denotes the number of unlabeled videos from VTW during training.}
\label{tab:semi}
\begin{tabular}{l|ccccc}
\hline
\# of unlabeled  & 0    & 500  & 1000 & 1500 & 2000  \\ \hline
SumMe  & 54.4 & 55.4 & 55.6 & 56.0 & 56.3  \\ \hline
TVSum  & 60.1 & 60.4 & 60.5 & 60.6 & 60.7  \\ \hline
\end{tabular}
\end{center}
\end{table}

\begin{table}[!t]
\begin{center}
\caption{Summary diversity Metric on SumMe and TVSum measured by Diversity Metric $\zeta$ ($\downarrow$), smaller $\zeta$ means higher diversity.}
\label{tab:diverse}
\begin{tabular}{c|cccc}
\hline
Database  &\scriptsize DPP-LSTM\cite{zhang2016video} &\scriptsize DR-DSN\cite{zhou2018deep} &\scriptsize SUM-GDA &\scriptsize SUM-GDA$_{unsup}$ \\\hline
SumMe & 0.133     & 0.109     & \textbf{0.099}    & 0.108              \\ \hline
TVSum & 0.319     & 0.312     & \textbf{0.304}    & 0.308              \\ \hline
\end{tabular}
\end{center}
\end{table}
\begin{figure*}[!t]
\begin{center}
\def\arraystretch{0.8}

\includegraphics[width=0.6\linewidth]{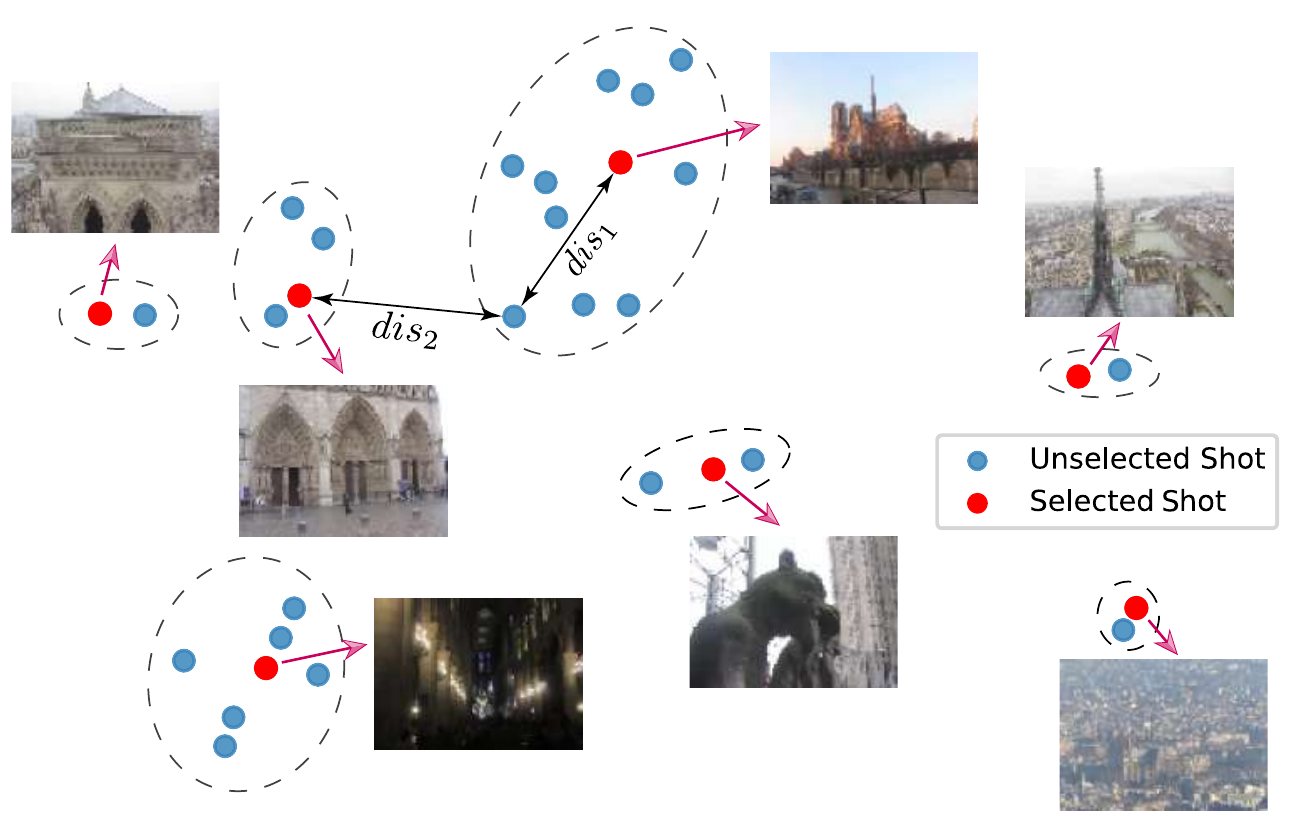}\\
{(a) Video 14 from SumMe: Notre Dame de Paris}\\
\vspace{0.2in}
\includegraphics[width=0.6\linewidth]{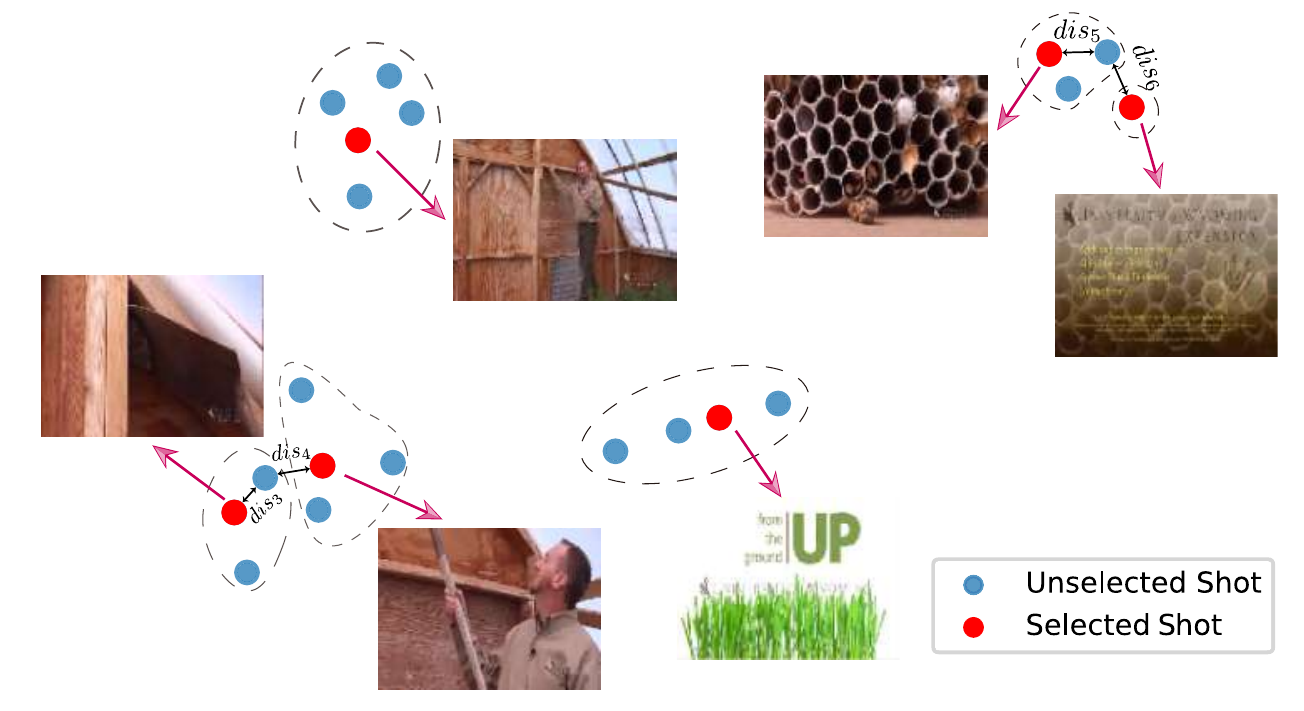}\\
{(b) Video 30 from TVSum: Paper Wasp Removal}
\end{center}
\caption{Diversity illustration using t-SNE \cite{maaten2008tsne} representation over the video shot features using SUM-GDA. The denser the group points (dashed circle), the more diversity the generated summaries. In (a), since $dis_1 < dis_2$, the blue circle belongs to the right group; similarly, in (b), $dis_3<dis_4$ and $dis_5<dis_6$, which decides the group the corresponding blue circle belongs to. Here $dis$ computes the distance between two circle points.}
\label{fig:t_sne}
\end{figure*}

\subsection{Diversity of Generated Summaries}
To evaluate diversity of summaries generated by different methods, we use the Diversity Metric ($\zeta$) defined in Eq.(\ref{diverse}), and the results are recorded in Table \ref{tab:diverse}. As the table shows, the proposed SUM-GDA approach has the lowest value $\zeta$, indicating it can generate the most diverse summaries on both of the two benchmark data sets. It can be attributed to the fact that SUM-GDA and SUM-GDA$_{unsup}$ can benefit from global diverse attention mechanism, which encourages the diversity of generated summaries.

Note that lower $\zeta$ means the video shots are closer to the corresponding key shot, \ie, the shots are more densely distributed. To illustrate the summary diversity metric $\zeta$, here we examine our method on two randomly selected videos from SumMe and TVSum respectively. The video shot distribution results are depicted in Figure~\ref{fig:t_sne}, which are obtained by using t-SNE \cite{maaten2008tsne} to project all the shot features into two-dimensional data space after inputting the given video to the SUM-GDA model. In this figure, the red solid circle denotes feature vector of key shot, blue solid circle denotes feature vector of unselected shot, and the dashed irregular circle denote the group including one key shot and its surrounding shots which are the closest ones.

From Figure~\ref{fig:t_sne}, it is vividly shown that with regard to different video shot groups (dashed circles), our model tends to select those shots leading to smaller distances with their neighbors (\ie~unselected shots) as the key shots to form the final summary. Besides, we can observe that the selected key shot and its group neighbors are densely clustered, which indicates the small $\zeta$ value according to the definition in Sec.~\ref{diversity}. Actually, the selected key shot acts as a representative shot in the group and may be linearly reconstructed from its group neighbors. Thus, the selected key shots from different groups are not only far away but also visually diverse. In another word, when the group shot points obtained by using the model are closer, the generated summaries will be more diverse.

\begin{table*}[t]
\begin{center}
\caption{Performance comparison with different input features in terms of Precision, Recall, F-score(\%).}
\label{tab:rgb_flow}
\begin{tabular}{l|c|c|c|c|c|c}
\hline
Database               & Method                             & RGB          & \small{Opt.Flow} & Precision     & Recall        & F-score       \\ \hline
\multirow{6}{*}{SumMe} & \multirow{3}{*}{SUM-GDA}           & $\checkmark$ &              & \textbf{52.5} & \textbf{53.6} & \textbf{52.8} \\
                       &                                    &              & $\checkmark$ & 46.9          & 49.8          & 47.9          \\
                       &                                    & $\checkmark$ & $\checkmark$ & 47.4          & 50.0          & 48.2          \\ \cline{2-7}
                       & \multirow{3}{*}{SUM-GDA$_{unsup}$} & $\checkmark$ &              & \textbf{49.6} & \textbf{52.3} & \textbf{50.0} \\
                       &                                    &              & $\checkmark$ & 47.6          & 52.0          & 49.8          \\
                       &                                    & $\checkmark$ & $\checkmark$ & 48.2          & 51.6          & 49.4          \\ \hline
\multirow{6}{*}{TVSum} & \multirow{3}{*}{SUM-GDA}           & $\checkmark$ &              & 59.0          & 58.9          & 58.9          \\
                       &                                    &              & $\checkmark$ & 59.2          & 59.3          & 59.2          \\
                       &                                    & $\checkmark$ & $\checkmark$ & \textbf{61.0} & \textbf{60.9} & \textbf{61.0} \\ \cline{2-7}
                       & \multirow{3}{*}{SUM-GDA$_{unsup}$} & $\checkmark$ &              & 59.5          & 59.7          & 59.6          \\
                       &                                    &              & $\checkmark$ & 59.4          & 59.1          & 59.2          \\
                       &                                    & $\checkmark$ & $\checkmark$ & \textbf{60.2} & \textbf{60.3} & \textbf{60.2} \\ \hline
\multirow{6}{*}{VTW}   & \multirow{3}{*}{SUM-GDA}           & $\checkmark$ &              & \textbf{50.1} & \textbf{50.7} & \textbf{50.2} \\
                       &                                    &              & $\checkmark$ & 41.7          & 49.5          & 43.3          \\
                       &                                    & $\checkmark$ & $\checkmark$ & 40.9          & 47.4          & 42.1          \\ \cline{2-7}
                       & \multirow{3}{*}{SUM-GDA$_{unsup}$} & $\checkmark$ &              & \textbf{47.9} & \textbf{48.6} & \textbf{47.9} \\
                       &                                    &              & $\checkmark$ & 37.2          & 42.1          & 38.1          \\
                       &                                    & $\checkmark$ & $\checkmark$ & 38.1          & 43.8          & 39.2          \\ \hline
\end{tabular}
\end{center}
\end{table*}

\subsection{Optical Flow Features}
Existing methods often use RGB images as the video input, which do not consider motion information within the video. To capture the video motion, we investigate the influences of optical flow features on generated summaries. Here we extract optical flow features of each video using TV-$L ^1$\cite{zach2007tvl1} package and we use the $pool 5$ features of GoogLeNet for fairness. We conduct the experiments on three data sets with three kinds of different inputs including RGB, Optical Flow, and the combination of the two. The experimental setting follows the Canonical setting described in Table~\ref{table:setting}. The performance results are shown in Table~\ref{tab:rgb_flow}.

From Table \ref{tab:rgb_flow}, we can observe that optical flow features greatly improve the performance on TVSum. This is because videos in this data set mostly contain a large portion of actions (\eg~motorcycle flipping, daily life of one woman) and the motion information plays an important role in learning the video summarization model. However, for SumMe and VTW data sets, the performances drop down when incorporating optical flow features. This might be for the reason that videos in these two data sets contain many different scenarios with less actions or slow-moved actions, but optical flow mainly captures the motion information while neglecting the still backgrounds which are also essential for learning the effective model. Sometimes, optical flow feature will even provide misleading direction to capture the motion information of video in such situation, resulting in generating less promising summaries.
%

\section{Conclusion and Future Work}
This paper has proposed a novel video summarization model called SUM-GDA, which exploits the global diverse attention mechanism to model pairwise temporal relations among video frames. In a global perspective, the mutual relations among different frame pairs in videos can be sufficiently leveraged for better obtaining informative key frames. To select the optimal subset of key frames, our model adopts determinantal point process to enhance the diversity of chosen frames. Particularly, determinantal point process results in different frame groups revealing diversified semantics in videos. Those chosen frames, indicated by high frame scores, are regards as the concise collection of source video with good completeness and least redundancy. Moreover, we extend SUM-GDA to the unsupervised scenario where the heavy cost of human labeling can be saved, and this can facilitate a variety of practical tasks with no supervised information. To investigate the summarization performance of the proposed models, we conducted comprehensive experiments on three publicly available video databases, \ie, SumMe, TVSum, and VTW. Empirical results have verified that both SUM-GDA and SUM-GDA$_{unsup}$ can yield more promising video summaries compared with several state-of-the-art approaches. In addition, we have examined the diversity of generated summaries and visualized the results, which suggest global diverse attention mechanism indeed benefits a lot in diversifying the key frames chosen for constituting video summary. Also the idea of global diverse attention might be helpful in training base learners for deep ensemble learning.

While we test our model on optical flow features of video which provides temporal dynamics, it does not always promote the performance and sometimes even reduce the summarization quality. Since temporal information is critical for encoding sequential data, it is worth putting more emphasis on improving the way of extracting optical flow from video and also exploring other possible ways to better capture temporal structure of video data in future. On the other hand, our model is advantageous in efficiency when the sequence length is less than the dimension of embedding representation, which is true mostly as frame sampling is adopted as preprocessing. However, this might not be the case when denser sampling is required in some situation which needs large computations. Therefore, it is a promising direction to further boost the efficiency of summarization model so as to make it be adapted to wider scenarios.
%


\bibliographystyle{IEEEtran}
\bibliography{pr_sumgda}

\ifCLASSOPTIONcaptionsoff
  \newpage
\fi

\end{document}